\pdfoutput=1

\documentclass[11pt]{article}

\usepackage[]{ACL2023}

\usepackage{times}
\usepackage{latexsym}
\usepackage{amssymb}
\usepackage{graphicx}
\usepackage{amsmath}
\usepackage{algorithm}
\usepackage{algpseudocode}
\usepackage{booktabs} 
\usepackage{xcolor}
\usepackage{todonotes}
\usepackage{array}
\usepackage{subcaption}
\usepackage[T1]{fontenc}
\usepackage{xspace}

\usepackage[utf8]{inputenc}

\usepackage{microtype}
\usepackage{amsmath}
\usepackage{inconsolata}
\usepackage{enumitem}
\usepackage{xcolor}
\definecolor{mybgcolor}{HTML}{ea998f} 
\definecolor{mybgcolor2}{HTML}{c4d1f7}
\definecolor{mybgcolor3}{HTML}{ffecc1}
\definecolor{mybgcolor4}{HTML}{aec5a8}

\title{Prototypical Reward Network for Data-Efficient RLHF}

\author{
    Jinghan Zhang\textsuperscript{\rm 1}, 
    Xiting Wang\textsuperscript{\rm 2*}, 
    Yiqiao Jin\textsuperscript{\rm 3}, 
    Changyu Chen\textsuperscript{\rm 2}, 
    Xinhao Zhang\textsuperscript{\rm 1}, 
    Kunpeng Liu\textsuperscript{\rm 1*} \\
  \textsuperscript{\rm 1}Portland State University, 
  \textsuperscript{\rm 2}Renmin University of China, 
  \textsuperscript{\rm 3}Georgia Institute of Technology \\
  \texttt{
    \{jinghanz,xinhaoz,kunpeng\}@pdx.edu} \\ 
    \texttt{\{xitingwang,chen.changyu\}@ruc.edu.cn} \\
    \texttt{yjin328@gatech.edu} \\
}

\newcommand{\model}{Proto-RM\xspace}
\newcommand{\modelbf}{\textbf{Proto-RM}\xspace}

\newcommand\blfootnote[1]{%
  \begingroup
  \renewcommand\thefootnote{}\footnote{#1}%
  \addtocounter{footnote}{-1}%
  \endgroup
}

\definecolor{titlecolor}{HTML}{D8EAD2} 

\begin{document}
\maketitle
\begin{abstract}
The reward model for Reinforcement Learning from Human Feedback (RLHF) has proven effective in fine-tuning Large Language Models (LLMs). Notably, collecting human feedback for RLHF can be resource-intensive and lead to scalability issues for LLMs and complex tasks. 
Our proposed framework \modelbf leverages prototypical networks to enhance reward models under limited human feedback. By enabling stable and reliable structural learning from fewer samples, \model significantly enhances LLMs' adaptability and accuracy in interpreting human preferences. Extensive experiments on various datasets demonstrate that \model significantly improves the performance of reward models and LLMs in human feedback tasks, achieving comparable and usually better results than traditional methods, while requiring significantly less data. in data-limited scenarios. 
This research offers a promising direction for enhancing the efficiency of reward models and optimizing the fine-tuning of language models under restricted feedback conditions. 
\end{abstract}

\section{Introduction}

\par Reinforcement Learning from Human Feedback (RLHF) effectively integrates meticulous human judgment with the model's capacity for large-scale data processing~\cite{cortes2015advances,bai2022training,stiennon2020learning}, enhancing language models' adaptability to human communication styles and preferences~\cite{yuan2023rrhf}. 
By utilizing Reinforcement Learning (RL) over supervised fine-tuning, RLHF captures the complexity of human language, involving emotions, context, and subtle linguistic differences~\cite{ouyang2022training,wang2023densecl}
thereby offering enhanced adaptability and flexibility in human interactions. \blfootnote{\textsuperscript{*}Corresponding Authors}

\par The success of RLHF hinges on the quality of the reward model~\cite{wang2024secrets,lee2023rlaif,bai2022constitutional,gilardi2023chatgpt}, which guides the RLHF learning process, ensures its accuracy and efficiency~\cite{ouyang2022training}, while preventing deviations from desired outcomes~\cite{paulus2017deep}. 
A deficient reward model, however, may learn a complex yet inaccurate error surface, leading the model to favor high-scoring yet erroneous solutions 
~\cite{chen2019generative,chen2023semi,li2017deep,yang2024foundation}. This overfitting can result in responses that maximize rewards but diverge from the actual objectives and human preferences~\cite{10.1145/3447548.3467105}. 
Notably, effective RLHF typically requires extensive data~\cite{sun2023exploring,zhang2024ratt,zhang2024tfwt}. 
\paragraph{This Work.} To address these challenges, we integrate prototypical networks--instance-based algorithms that learn representative prototypes for similar examples to facilitate tasks like classifications or regression~\cite{snell2017prototypical}--with the reward model for RLHF. 
Prototypical networks are particularly suitable for few-shot learning as they efficiently extract key features from limited samples for decision-making~\cite{liu2020prototype}. 
By optimizing the embedding process in the reward model using prototypical networks, we enable the reward model to learn stable and reliable data representation structures with limited sample sizes. 
This method is especially suitable in enhancing the model's learning and generalization from human feedback samples, given the limitations of sample quantity and the complexity of human preferences~\cite{bai2022constitutional}.

\begin{figure*}
  \centering
  \includegraphics[width=6.2in]{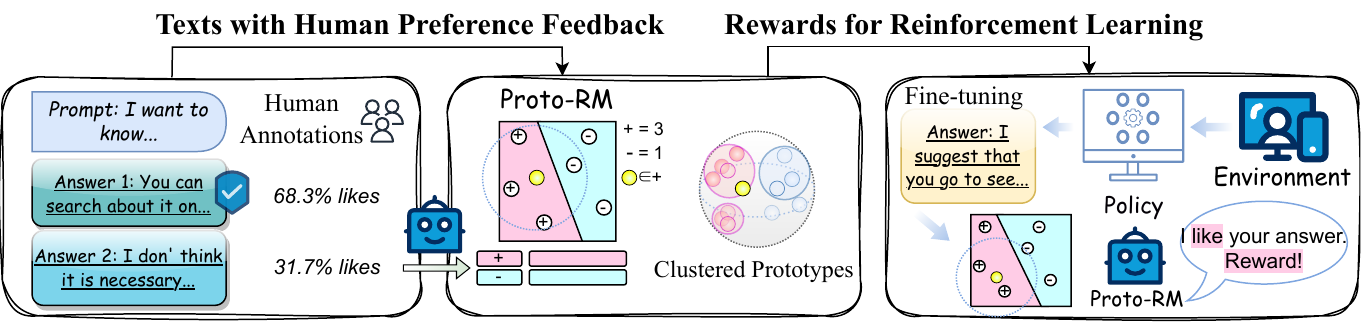}
  \caption{Our proposed \model framework enhances reward model with prototypical networks. \textbf{Left}: Humans annotate pairwise RLHF data and select their preferred text. \textbf{Middle}: \model aggregates similar examples from embedding space into prototypes. \textbf{Right}: The enhanced reward model fine-tunes a pretrained LLM.}
  

  \label{fig:intro}
  \vspace{-2mm}
\end{figure*}
\par To enhance the effectiveness of the reward model with limited human feedback data, we propose the \underline{\textbf{Proto}}typical \underline{\textbf{R}}eward \underline{\textbf{M}}odel (\modelbf) that decreases reliance on human feedback without compromising the performance of the reward model. The fundamental principle of the reward model is to assimilate human feedback to evaluate and steer the model outputs to meet human expectations and standards. Its crucial capability lies in effectively learning and extracting vital parameter information from limited human feedback, thus guiding the model's behavior. 
Our approach focus on learning a method that performs well in few-shot scenarios and is suitable for learning from human feedback samples while preserving the reward model's network structure and enhancing the parameterization capabilities. 

\par Our method consists of three pivotal steps. First, in \textbf{Sample Encoding and Prototype Initialization}, 
we employ a reward model to encode samples. These encodings serve as the basis for initializing prototypes with a strategically selected subset of the encoded samples. Subsequently, we analyze the relationship between these initialized prototypes and the encodings of other samples. 
Second, in \textbf{Prototype Update and Addition}, we continuously refine the sample encodings based on their distances to the prototypes. Concurrently, we update the reward model's parameters by validating the predictions generated through the refined encodings. This iterative refinement ensures that prototypes accurately mirror the attributes of samples, enhancing the learning efficacy through human feedback samples. 
Finally, \textbf{Reward Model Fine-tuning} employs the refined prototypes and encodings 
to train the reward model. This training aims to accurately evaluate and guide the outputs of the language model, thereby improving the performance of LLMs during the fine-tuning stage. 

\paragraph{Contributions.} Our main contributions include:
\begin{itemize}[leftmargin=0.12in]
    \item We propose \model, a novel prototypical-network-based method to improve the reward model. This structure facilitates training with fewer human feedback samples without compromising the learning ability of the reward model in scenarios with ample samples.
    
    \item We explore a prototypical learning method for human feedback samples, effectively managing human feedback that is difficult to quantify and varies in length.
    
    \item We conduct a series of experiments to validate the effectiveness and robustness of our method \model across different dataset sizes and evaluate the performance of LLM fine-tuned by \model. Our experiments show that our method has clear advantages and achieves the effectiveness of training with extensive data, even when using limited samples.
\end{itemize}

\section{Related Work}
\subsection{Reinforcement Learning from Human Feedback (RLHF)}
Large Language Models (LLMs) such as GPT-4~\cite{achiam2023gpt}, Bard~\cite{singh2023chat}, and LLaMA-2~\cite{touvron2023llama} have demonstrated significant capabilities in understanding human languages and preferences~\cite{zhao2023competeai,jin2023better, jin2024mm}. The efficacy of these models primarily hinges on Reinforcement Learning from Human Feedback (RLHF)~\cite{christiano2017deep,ziegler2019fine,ouyang2022training,casper2023open}, which enables LLMs to iteratively refine their text generation capabilities, thereby producing outputs that accurately reflect human values and preferences~\cite{song2023preference,wang2021reinforcing}. 
RLHF involves three interconnected stages: feedback collection, reward modeling, and policy optimization. Initially, human annotators assess model outputs and provide feedback. This feedback is then utilized in the reward modeling process, where supervised learning is used to train a reward model that replicates human assessments~\cite{lambert2023history,dong2019unified}. Finally, during policy optimization, the model is fine-tuned to produce outputs that receive positive evaluations from the reward model~\cite{zheng2023secrets}. RLHF excels at identifying ``good'' behaviors than other reward specification or learning methods. However, it faces significant challenges due to the large volume of human feedback data required, which can be costly and resource-intensive~\cite{beeching2023stackllama,yao2023instructions}.

\subsection{Prototypical Networks}
Prototypical Learning is a powerful approach for improving model interpretability and accuracy in few-shot classification scenarios~\cite{liu2020prototype,kim2014bayesian}. Researchers have extensively enhanced prototypical networks for category learning~\cite{pan2019transferrable,ding2020graph,ji2020improved}. The advantages of prototypical networks lie in their simplicity and intuitiveness, enabling rapid adaptation to new samples and categories without the need for extensive data or complex training processes~\cite{fort2017gaussian,yao2023value}. While these networks are commonly used in classification problems with distinct category labels, their application is notably absent in the domain of non-quantitative semantic understanding and text comparison.
\section{Problem Formulation}
The primary objective is to train a reward model capable of training a policy that generates high-quality texts, as evaluated by humans, using a constrained set of human-annotated data. 

\paragraph{Input.} The input of the reward model is a dataset $\mathcal{D} = \left\{(x_i, y^{+}_i, z^{+}_i, y^{-}_i,z^{-}_i) \right\}_{i=1}^N$ with $N$ examples, where each example consists of a common post $x_i \in \mathbf{X}$ and two corresponding summaries $y^{+}, y^{-} \in \mathbf{Y}$. These summaries are distinguished by human preferences: $y^{+}$ is the preferred (\textit{chosen}) response with annotation $z_i^{+} \in \mathbf{Z}$, and $y^{-}$ is the less preferred (\textit{rejected}) response 
with annotation $z_i^{-} \in \mathbf{Z}$, where $\mathbf{Z} = \left\{\text{chosen}, \text{rejected}\right\}$. 
\paragraph{Output.} The outputs consist of 2 components: 
\begin{itemize}[leftmargin=0.12in]
\vspace{-2mm}
    \item Predictive scores $s_{(x_i, y^{+}_i)}$ and $s_{(x_i, y^{-}_i)}$ for each example $(x_i, y_i^{+}, y_i^{-})$, indicating the model's assessment of the relative quality of the summaries;
    \item The reward model $f_{\phi}: \mathbf{X} \times \mathbf{Y} \rightarrow \mathcal{E}$, where $\mathcal{E}$ is the embedding space. $f_{\phi}$ includes an embedding function $e_{\phi}$ and an aligned linear scoring process.
\end{itemize}
\section{Methodology}
The primary objective is to train a reward model that predicts which answer $\{y_i^{+}, y_i^{-}\}$ is better according to human judgment, given a prompt $x_i$. 

\subsection{Reward Model with Prototypical Network}
\paragraph{Reward Model for RLHF.} The reward model evaluates the quality of outputs generated by the language model. The model's feedback guides fine-tuning so that the model outputs align with human preferences. Given the input dataset $\mathcal{D}$, the RLHF reward model, parameterized by $\phi$,  converts text pairs into encodings in the embedding space $\mathcal{E}$: 
\begin{equation}
\mathbf{f}_{\phi}(x, y) \rightarrow \mathbf{e} \in \mathcal{E}, \mathbf{e} = (\mathbf{e}_x, \mathbf{e}_y).
\end{equation}
Here, $\mathbf{e}$ is the representation of the input pair $(x, y)$, $\mathbf{e}_x$ and $\mathbf{e}_y$ are the representations of the prompt and answer, respectively. 

\paragraph{Prototypical Network.} In the prototypical network, a set of prototype vectors $ \mathbf{p}_k$ is categorized into two groups: $ \mathbf{p}^{+} $ and $ \mathbf{p}^{-} $. The classification of each sample pair's embedding $ \mathbf{e}_{(x_i, y^{*}_i)}$, where $y^{*} \in \{y^{+}, y^{-}\}$, is determined by the proportion of these two classes of prototypes within the adjacent prototypes. The embedding $ \mathbf{e}_{(x_i, y^{*}_i)} $ is updated based on all the prototype vectors in their respective category, with weights assigned according to their importance. The importance of prototype $\mathbf{p}_k$ is computed using the distance metric $ d(\cdot, \cdot) $:
\begin{equation}
    \mathbb{P}(\mathbf{p}_k | (x_i, y^{*}_i)) \propto \exp(-d(\mathbf{e}_{(x_i, y^{*}_i)}, \mathbf{p}_k)),
\end{equation}
\noindent where $ d(\cdot, \cdot) $ is usually taken as squared L2 distance.
We then update the embedding for each sample according to its class. For a sample embedding related to the $ \mathbf{p}^{*} $ prototype, we update its embedding $ \mathbf{e}_{(x_i, y^{*}_i)} $ using all $|\mathbf{P}|$  prototypes in class $ \mathbf{p}^{*} $, where $|\mathbf{P}|$ is the total number of prototypes of class `$*$'. 
The formula for updating the embedding is expressed as:
\begin{equation}
  \mathbf{e}_{(x_i, y^{*}_i)} = \frac{1}{|\mathbf{P}|} \sum_{k=1}^{|\mathbf{P}|} (\mathbb{P}(\mathbf{p}_k | (x_i, y^{*}_i)) \cdot \mathbf{p}_{k}) .
  \label{eq:update embedding}
\end{equation}
The updated embedding is then transformed into a score within a linear layer.
\begin{figure*}
  \centering
  \includegraphics[width=\textwidth]{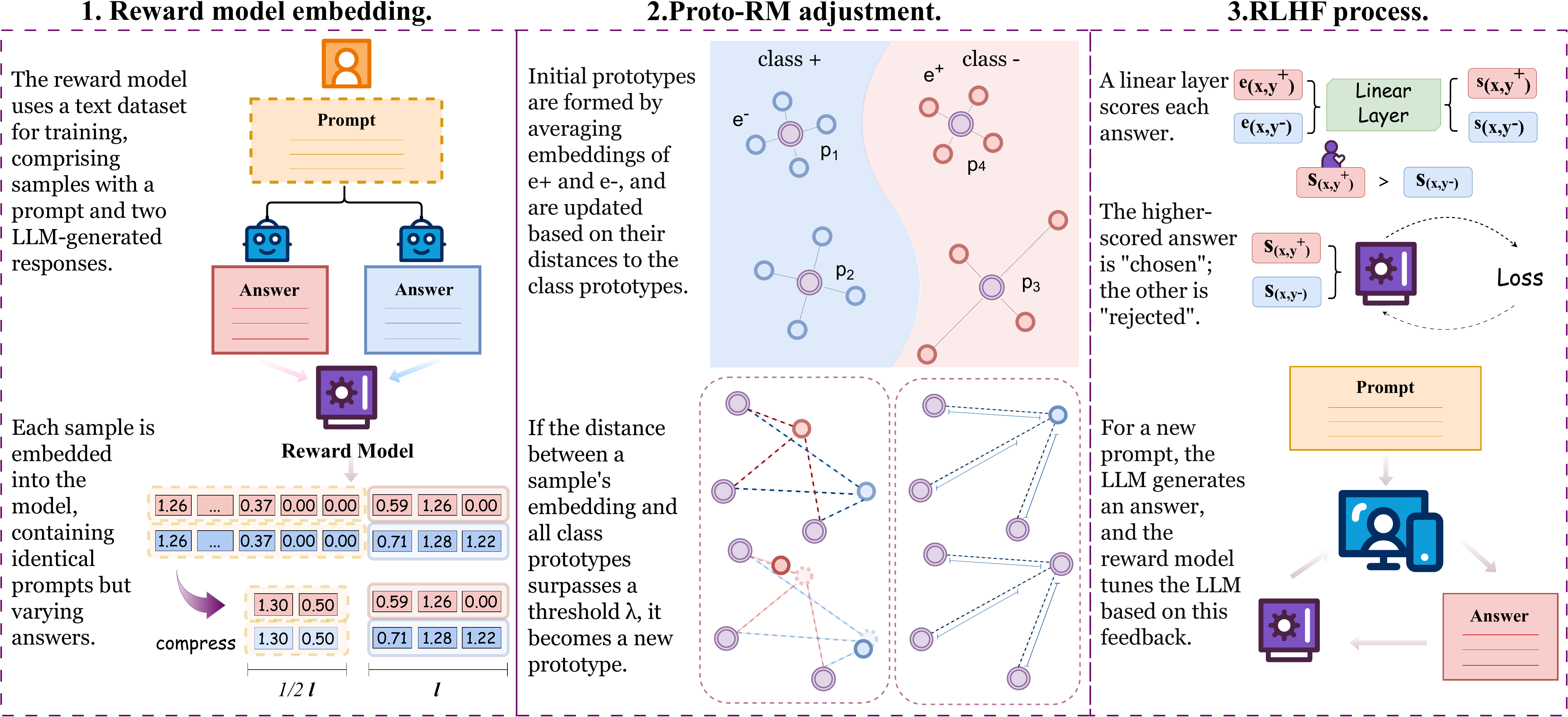}
  \vspace{-6mm}
  \caption{
The framework consists of three components: 1) Reward model embedding, 2) Proto-RM adjustment and 3) RLHF process. The reward model compress and align the sample text pair embeddings to produce representative prototypes, and the prototypes adjust the embeddings to update the reward model. 
  }
  \label{fig:framework}
  \vspace{-5mm}
\end{figure*}

\subsection{Reward Model with Prototypical Network} \paragraph{Prototype Initialization.} During the initialization phase, our goal is to properly initialize two classes of prototypes $ \mathbf{p}_k \in \{\mathbf{p}^+, \mathbf{p}^-\} $. We randomly select $ n $ sample pairs and aggregate them according to their labels $ z_i $. Specifically, we initialize different prototypes using the sample embeddings labeled ``chosen'' and ``rejected''. This strategy is employed to allow the model to better learn human preferences instead of mere differences in the content of the samples.
We process the prompt and answer components of each text bar separately. For embeddings that initialize $\mathbf{p}_0$ to be the original prototypes
, we perform pairwise sample alignment to ensure uniformity and fairness in compression and computation across positive and negative examples. This alignment method guarantees that the prototypes are updated consistently, reflecting a balanced representation of both prompt and answer components in the embedding space: 
\begin{equation}
    \mathbf{e}_{x_i} \leftarrow  \operatorname{align}(\mathbf{e}_{x_i}, \max \|\mathbf{e}_{x_i}\|),
    \label{eq:align}
\end{equation}
\noindent where $\operatorname{align}(\cdot)$ denotes updating the embedding $\mathbf{e}_{x_i}$ to a new vector with the same maximum length as the longest embedding vector among all $\mathbf{e}_{x_i}$. Elements beyond the original length of $\mathbf{e}_{x_i}$ are padded with zeros. Similarly, we have:
\begin{eqnarray}
    &\mathbf{e}_{y^{*}_i} \leftarrow \operatorname{align}(\mathbf{e}_{y^{*}_i}, \max \|\mathbf{e}_{y^{*}_i}\|), \\
    &\mathbf{e}_{(x_i,y^{*}_i)} = (\mathbf{e}_{x_i}, \mathbf{e}_{y^{*}_i}).
\end{eqnarray}

An initial prototype constructed from $ n $ text pairs is defined as $ \mathbf{p}_0 = \frac{1}{n} \sum \mathbf{e}_{(x_i, y^{*}_i)}$, with a length of $ \|\mathbf{p}_0\| = (\max \|\mathbf{e}_{x_i}\| + \max \|\mathbf{e}_{y^{*}_i}\|), i = 1, 2, \ldots n $. This ensures that the prototype encapsulates the essential features of both prompt and answer.

We derive an initial set of $K$ prototypes using mean pooling as the aggregate function while keeping the parameters of the reward model $\phi$ frozen. 
During initialization, we disable gradient updates of the prototype vectors, ensuring that other model parameters do not affect the initialization process and thus guaranteeing its robustness. 



\paragraph{Prototype Update.} Our goal is to represent the samples effectively and comprehensively using prototypes. However, a fixed number of prototypes may not suffice for this purpose. Too few prototypes can lose important information, while too many can affect their representativeness and increase computational costs~\cite{snell2017prototypical,ming2019interpretable,jin2022prototypical}. 
Therefore, we consider employing Infinite Mixture Prototypes (IMP)~\cite{allen2019infinite} to automatically generate prototypes during training. This approach allows the model to increase the number of prototypes as needed, based on the distance relationship between the prototypes and the samples. The IMP technique is commonly used in classifying graphical samples, but its application in textual information is relatively less frequent. Prototype methods excel in processing graphical samples with their visual and intuitive features, but text's abstract and multidimensional characteristics, covering semantics, syntax, and context, complicate their use in textual data. Due to our reliable embedding and alignment of text samples using the reward model, we successfully implement IMP for effective learning from human feedback samples.

After initializing the prototypes, we use the prototypical network to better assimilate new inputs. To enhance the representativeness and diversity of the prototypes, we 1) appropriately generate new prototypes and 2) continually update existing ones.

\paragraph{Generating New Prototypes.} We define the set of prototypes as $\mathbf{P}$. To increase the representativeness and diversity of the prototypes, for each sample $(x_i, y^{*}_i) \in \mathcal{D}$, if the minimum distance between $\mathbf{e}_{(x_i, y^{*}_i)}$ and any prototypes in $\mathbf{P}$ exceeds a threshold $\lambda$, we create a new prototype based on $\mathbf{e}_{(x_i, y^{*}_i)}$. The threshold distance $\lambda$ is defined as:
\begin{equation}
\lambda = 2\sigma \log \left( \frac{\alpha}{(1 + \frac{\rho}{\sigma})^{d/2}} \right) ,
\label{eq:imp_lambda}
\end{equation}
where $\sigma$ is the cluster variance learned jointly with $\phi$, $\rho$ is the standard deviation for the base distribution from which the cluster means are sampled, and $\alpha$ is a hyperparameter controlling the concentration of clusters in the Chinese Restaurant Process~\cite{wang2009variational}. Our approach can balance between fitting simple data distributions with low capacity and complex distributions with high capacity.

We then compute the distance from each text bar in a text pair to every prototype $\mathbf{p}_k$ in their class, denoted as $d(\mathbf{e}_{(x_i, y^{*}_i)}, \mathbf{p}_k)$. Using the negative of these distances, we calculate the softmax to obtain the probability distribution of sample $(x_i, y^{}_i)$ belonging to prototype $\mathbf{p}_j$. 
Additionally, during the update of sample embeddings, we incorporate a proportionate dropout of the prototypes, which enhances the model's ability to generalize and avoid overfitting to specific patterns:

\vspace{-5mm}
{\small
\begin{equation}
P( \mathbf{p}_i= \mathbf{p}_j | (x_i, y^{*}_i)) = \frac{\exp(-d(\mathbf{e}_{(x_i, y^{*}_i)}, \mathbf{p}_j))}{\sum_{k=1}^{\lfloor \rho K \rfloor} \exp(-d(\mathbf{e}_{(x_i, y^{*}_i)}, \mathbf{p}_k))},
\end{equation}}
\noindent where $\rho$ is the dropout ratio, $ K $ is the total number of prototypes, and $\lfloor \cdot \rfloor$ represents the floor function. Instead of random dropout, \model calculate the cosine similarity of prototypes within the same class and drop prototypes with the highest similarity. This approach ensures that the remaining prototypes are more diverse and thus more representative of the data.  
The new embedding $\mathbf{e}'_{(x_i, y^{*}_i)}$ is derived using the weighted average of $\mathbf{p}_k$ with respect to the probability distribution: 
\begin{equation}
\mathbf{e}'_{(x_i, y^{*}_i)} = \sum_{k=1}^K P(\mathbf{p}_i = \mathbf{p}_k | (x_i, y^{*}_i)) \cdot \mathbf{p}_k.
\end{equation}

\paragraph{Annotation Prediction.} We then evaluate the performance of the model and update it. We predict the annotation $z_i$ of the new embedding $ \mathbf{e}'_{(x_i, y^{*}_i)} $. The embedding transform into a score $ s_{(x_i, y^{*}_i)} $ through a linear layer. By comparing the scores $ s_{(x_i, y^{+}_i)} $ with $ s_{(x_i, y^{-}_i)} $, the model annotate the one with the higher score as ``chosen'', and the one with the lower score as ``rejected''. We evaluate the model's predictions $z_i$ against real human annotations and perform backpropagation accordingly.


\paragraph{Loss and Backpropagation.} The final step involves the computation of the overall loss, including reward loss and diversity loss to enhance the model's performance and reduce the risk of overfitting. Inspired by~\citet{stiennon2020learning}, we define the reward loss $\mathcal{L}_r$ as:
\begin{equation}
\begin{split}
\mathcal{L}_r = & -\mathbb{E}_{(x_i,y^{+}_i,y^{-}_i)\sim \mathcal{Z}}[  \log(\sigma(r_{\phi}(x_i, y^{+}_i) \\
& - r_{\phi}(x_i, y^{-}_i)))],
\end{split}
\end{equation}

\noindent where $r_{\phi}(x_i, y^{*}_i)$ is the scalar output of the reward model for prompt $x_i$ and answer $y^{*}_i$ with parameter $\phi$, and $\mathcal{Z}$ is the collection of human annotations. At the end of training, we normalize the reward model outputs such that the reference text pairs from the dataset achieve a mean score of 0.

For diversity loss $\mathcal{L}_{\text{div}}$, in order to ensure a sparse distribution among prototypes, we employ a hyperparameter $\tau$ to constrain their average in-between Euclidean distances. As model parameters, prototypes are involved in backpropagation through gradient descent, allowing for dynamic refinement. The sparsity constraint is implemented via a diversity loss $\mathcal{L}_{\text{div}}$~\cite{ji2022few}, which is guided by the average Euclidean distance between prototypes:

\begin{equation}
\psi = \begin{cases}
\text{Euc}(\Phi) - \tau & \text{if } \text{Euc}(\Phi) \geq \tau, \\
\tau - \text{Euc}(\Phi) & \text{if } \text{Euc}(\Phi) < \tau,
\end{cases}
\end{equation}
\begin{equation}
\mathcal{L}_{\text{div}} = \log(\psi + 1).
\end{equation}

The full objective $ \mathcal{L} $ linearly combines $ \mathcal{L}_r $ and $ \mathcal{L}_{\text{div}} $ using a hyperparameter $ \rho_d $:
\begin{equation}
\mathcal{L} = \mathcal{L}_r + \rho_d \mathcal{L}_{\text{div}} .
\end{equation}

\begin{algorithm}[t]
\caption{Reward Model with Prototypical Networks}
\begin{algorithmic}[1]
\State \textbf{Input}: $\mathcal{D} = \left\{ (x_i, y^{+}_i, y^{-}_i), (z^{+}_i,z^{-}_i) \right\}_{i=1}^N$, where each $z^{+}, z^{-} \in Z = \left\{\text{chosen}, \text{rejected}\right\}$
\State \textbf{Output}:  The predicting score pair $S(x, y^{+}, y^{-}) = (s^{+}, s^{-})$ and the reward model $f_{\phi}$
\State  Initialize $K$ Prototypes through \textbf{Prototype Initialization}
\For{minibatch $B_r \in \mathcal{D}$}
    \State \parbox[t]{\dimexpr\linewidth-\algorithmicindent}{Perform \textbf{Prototype Update and Addition} and estimate $\lambda$ according to Eq.~\ref{eq:imp_lambda}}
    \For{$(x_i, y_i^{+}, y_i^{-}) \in B_r$}
        \State Converts $(x_i, y_i^{+}, y_i^{-})$ into encodings 
        
        \Statex \qquad \quad $\mathbf{e}_{(x_i, y_i^{+})}$ and $\mathbf{e}_{(x_i, y_i^{-})}$

        \For{$y_i^{*} \in {y^{+},y^{-}}$}

            \State Allign $\mathbf{e}_{(x_i, y_i^{*})}$ according to Eq.~\ref{eq:align}
        
            \State {Calculate $d_{i,k} = d(\mathbf{e}_{(x_i, y^{*}_i)}, \mathbf{p}_k)$ for
            
            \Statex \qquad \: \qquad  $\mathbf{p}_k \in \mathbf{p}^*$, and $d_{i,k} = +\infty$ for $\mathbf{p}_k \notin$ 
            
            \Statex \qquad \: \qquad$ \mathbf{p}^*$}

            \State Update the embedding according  
            
            \Statex \qquad \: \qquad to Eq.~\ref{eq:update embedding}
            
            \If{$\min d_{i,k} > \lambda$}
                \State Create the $K + 1$-th prototype 
                \Statex \qquad \qquad \qquad $\mathbf{p}_{K+1}$ using $\mathbf{e}_{(x_i, y_i^{*})}$; Increment 
                \Statex \qquad
                \qquad \qquad $K$ by 1
            \EndIf
            \State Compute $s_{(x_i, y^{*}_i)}$ though \textbf{Annota-}
            
            \Statex \qquad \qquad \: \textbf{tion Prediction}

        \EndFor
    \EndFor
\EndFor
\end{algorithmic}
\end{algorithm}

\vspace{-5mm}
\section{Experiments}

In this section, we first compare the consistency of annotations between Proto-RM and Baseline Reward Model (Baseline RM) with real human feedback on Prompt-Answer text pairs. Subsequently, we contrast the differences in text quality of LLM outputs after fine-tuning with Proto-RM versus Baseline RM. Following this, we explore the significance of different modules in the learning of the reward model, assessing the effectiveness of our innovative points.
\subsection{Experiment Settings}
\paragraph{Datasets.} 
We train reward models using three datasets at varying data proportions. The datasets employed are as follows:

\begin{table}[htbp]
\centering
\small 
\begin{tabular}{crrr}
\toprule
& \textbf{Webgpt} & \textbf{Pairwise} & \textbf{Summarize} \\
\midrule
\textbf{5\%}   & 979    & 1,657  & 9,692   \\
\textbf{10\%}  & 1,958  & 3,314  & 19,384  \\
\textbf{20\%}  & 3,916  & 6,629  & 38,768  \\
\textbf{Total} & 19,578 & 33,143 & 193,841 \\
\bottomrule
\end{tabular}
\caption{Data distribution across different datasets.}
\vspace{-2mm}
\end{table}

\begin{itemize}[leftmargin=0.12in]
    \item \textit{Webgpt Comparisons} (Webgpt)~\cite{nakano2021webgpt} contains pairs of model answers with human preference scores in the WebGPT project. 
    \item \textit{Synthetic Instruct GPT-J Pairwise} (Pairwise)~\cite{alex-etal-2021-online} contains human feedback for reward modeling, featuring pairwise summary evaluations and Likert scale quality assessments. 
    \item \textit{Summarize from Feedback} (Summarize)~\cite{stiennon2020learning} contains pairwise summaries with human annotations from the TL;DR dataset.
\end{itemize}


\paragraph{Implementation Settings.} We use GPT-J~\cite{gpt-j} as the pre-trained LLM. Our experiment applies a batch size of 8 and initialize each prototype using $n = 2$ examples. The sequence length is set to $550$. We fix the value of $\alpha$ at $0.1$ and the initial value of $\rho$ at $5$. We use the AdamW optimizer~\cite{zhuang2022understanding} and search the best learning rate within the range of $[1e-6, 1e-5]$. Other hyperparameters are set to their default values as in \citet{allen2019infinite}. 
We use the trlX framework~\cite{havrilla-etal-2023-trlx} for model implementation.
All experiments are conducted for a maximum of $5$ epochs with early stopping on a server with NVIDIA Tesla A100 GPU (80GB memory).

\subsection{Comparison with Baseline Reward Model}
\begin{table*}[htbp]
\centering
\small 
\setlength\tabcolsep{4pt} 
\begin{tabular}{l|cccccc}
\hline
\textbf{Datasets} & \multicolumn{2}{c}{\textbf{Webgpt}} & \multicolumn{2}{c}{\textbf{Pairwise}} & \multicolumn{2}{c}{\textbf{Summarize}} \\
\hline
\textbf{RM} & \textbf{Baseline-RM} & \textbf{Proto-RM} & \textbf{Baseline-RM} & \textbf{Proto-RM} & \textbf{Baseline-RM} & \textbf{Proto-RM} \\
\hline
\textbf{5\%}   & $57.46 \pm 0.21$ & $\mathbf{58.94 \pm 0.22 \textcolor{blue}{(\raisebox{0.25ex}{$\scriptstyle{+}$}1.48)}}$ & $98.96 \pm 0.15$ & $\mathbf{99.44 \pm 0.18 \textcolor{blue}{(\raisebox{0.25ex}{$\scriptstyle{+}$}0.48)}}$ & $65.36 \pm 0.19$ & $\mathbf{67.67 \pm 0.23 \textcolor{blue}{(\raisebox{0.25ex}{$\scriptstyle{+}$}2.31)}}$ \\
\textbf{10\%}  & $58.86 \pm 0.24$ & $\mathbf{59.30 \pm 0.26 \textcolor{blue}{(\raisebox{0.25ex}{$\scriptstyle{+}$}0.44)}} $ & $99.14 \pm 0.17$ & $\mathbf{99.65 \pm 0.20 \textcolor{blue}{(\raisebox{0.25ex}{$\scriptstyle{+}$}0.51)}}$ & $66.51 \pm 0.21$ & $\mathbf{67.76 \pm 0.25\textcolor{blue}{(\raisebox{0.25ex}{$\scriptstyle{+}$}1.25)}}$ \\
\textbf{20\%}  & $58.41 \pm 0.28$ & $\mathbf{60.56 \pm 0.29\textcolor{blue}{(\raisebox{0.25ex}{$\scriptstyle{+}$}2.15)}}$ & $99.45 \pm 0.16$ & $\mathbf{99.84 \pm 0.11 \textcolor{blue}{(\raisebox{0.25ex}{$\scriptstyle{+}$}0.39)}}$ & $67.46 \pm 0.22$ & $\mathbf{68.72 \pm 0.27 \textcolor{blue}{(\raisebox{0.25ex}{$\scriptstyle{+}$}1.26)}}$ \\

\hline
\end{tabular}
\caption{Comparison of \model and Baseline across various datasets and sizes. Proto-RM consistently \textcolor{blue}{outperforms} Baseline-RM in terms of accuracy.}
\label{tab:RM compare}
\vspace{-2mm}
\end{table*}



\begin{figure*}
    \centering
    \begin{minipage}{.32\textwidth}
        \centering
        \includegraphics[width=1\linewidth]{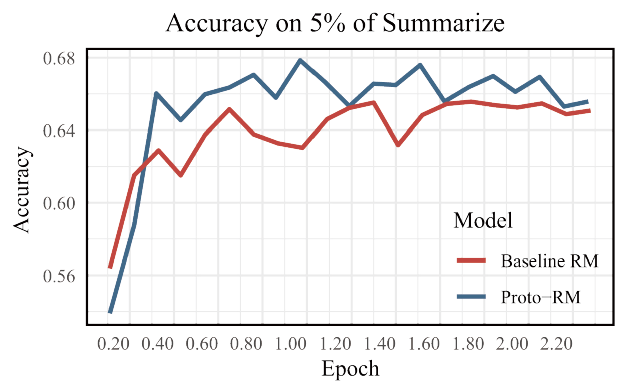}
    \end{minipage}%
    \begin{minipage}{.32\textwidth}
        \centering
        \includegraphics[width=1\linewidth]{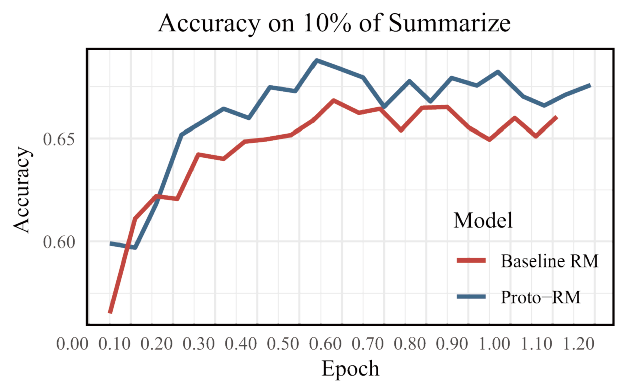}
    \end{minipage}
    \begin{minipage}{.32\textwidth}
        \centering
        \includegraphics[width=1\linewidth]{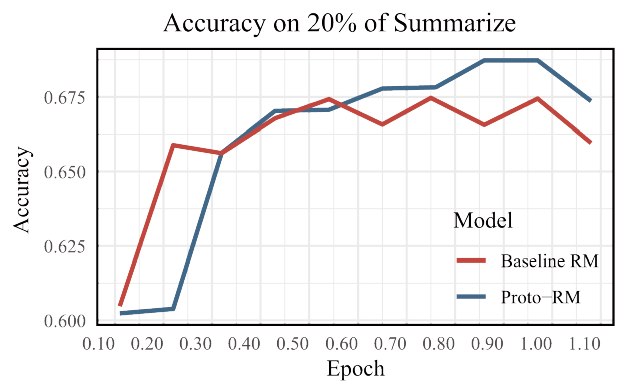}
    \end{minipage}
    \vspace{-2mm}
    \caption{Comparison of reward models' accuracy on 5\%, 10\%, and 20\% datasets.}
    \label{fig:combinedacc}
    \vspace{-3mm}
\end{figure*}

\begin{figure*}[h]
    \centering
    \begin{minipage}{.33\textwidth}
        \centering
        \includegraphics[width=1\linewidth]{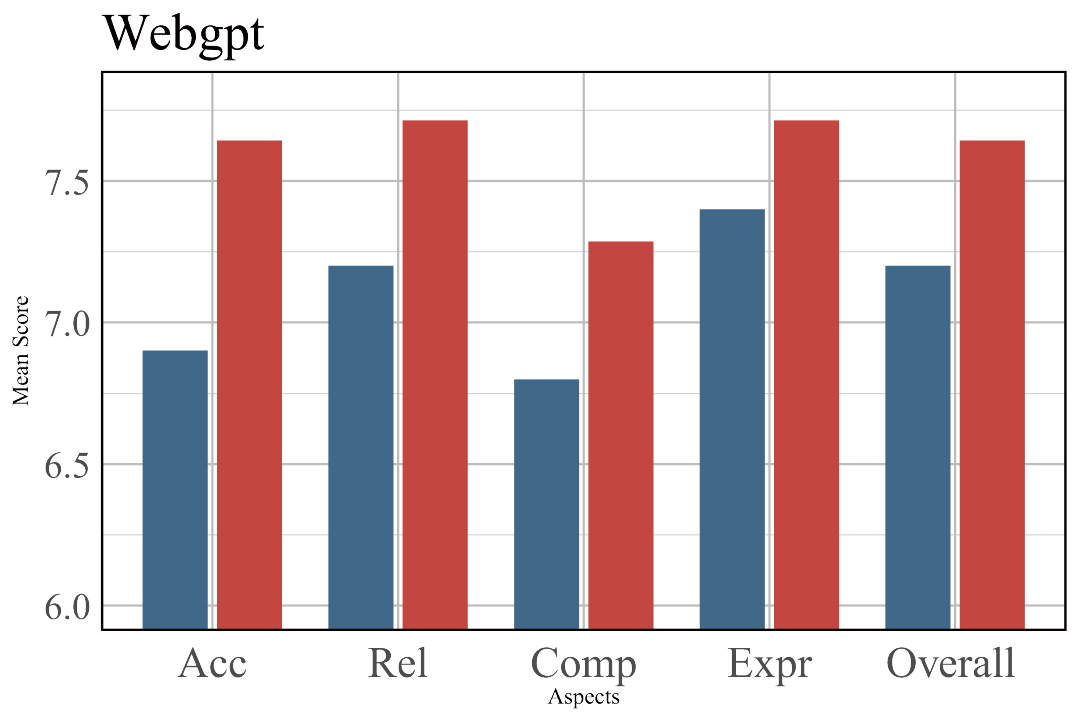}
    \end{minipage}%
    \begin{minipage}{.33\textwidth}
        \centering
        \includegraphics[width=1\linewidth]{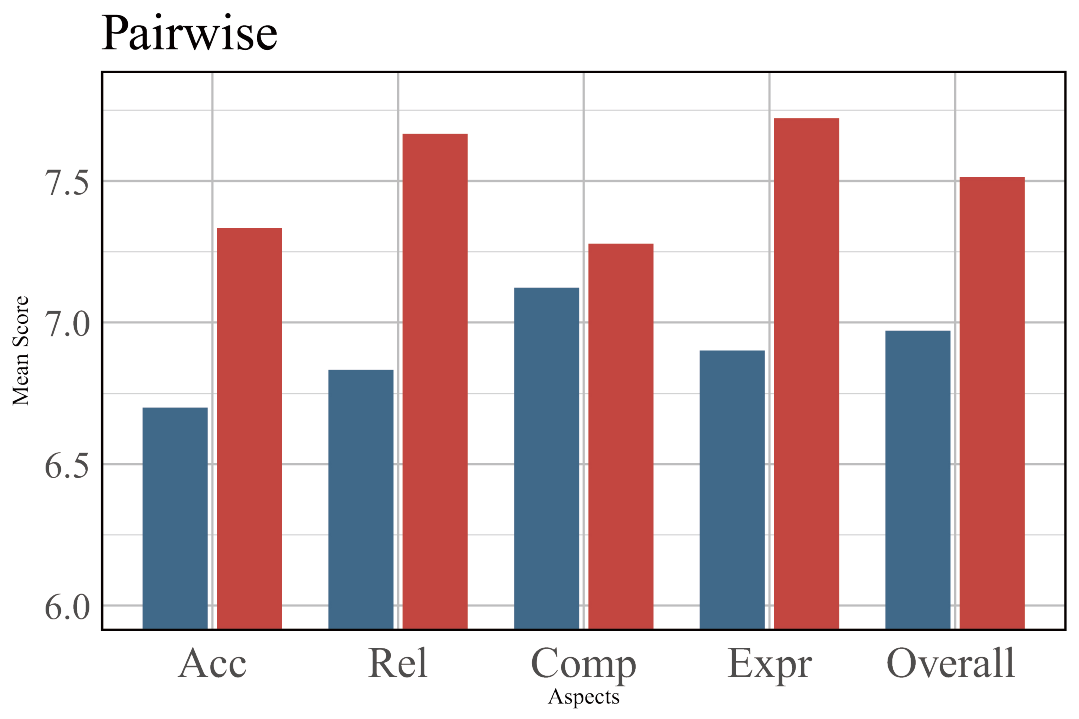}
    \end{minipage}
    \begin{minipage}{.33\textwidth}
        \centering
        \includegraphics[width=1\linewidth]{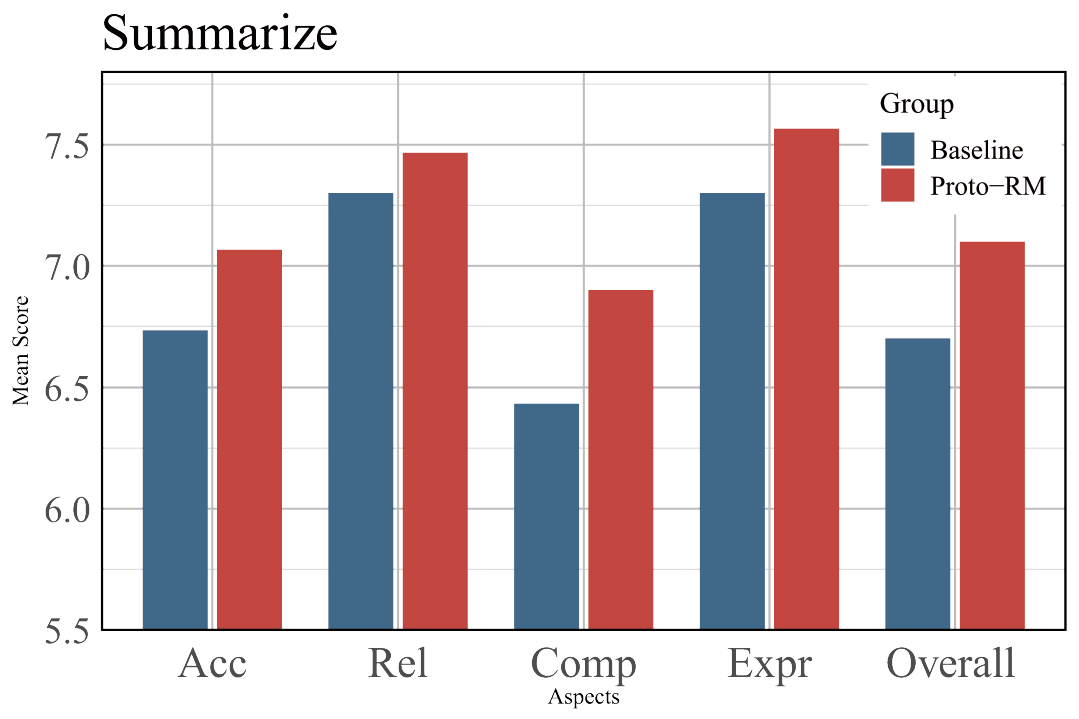}
    \end{minipage}
    \vspace{-2mm}
    \caption{Performance of LLM with reward model fine-tuning.}
    \vspace{-2mm}
    \label{fig:baselineVSours}
\end{figure*}
To compare the performance of Baseline RM and Proto-RM, we train and test both reward models on three datasets by different ratios. From Table~\ref{tab:RM compare} we can see that, across the different data proportions on the three datasets, Proto-RM consistently surpasses Baseline-RM. On the \textit{Webgpt} dataset, there is an accuracy improvement ranging from $1.48\%$ to $2.15\%$; on the \textit{Pairwise} dataset, the improvement spans from $0.48\%$ to $0.59\%$, with Proto-RM nearly achieving perfect accuracy; and on the \textit{Summarize} dataset, especially at the $20\%$ data proportion, Proto-RM exhibits the most significant accuracy gain of $1.26\%$.
\par The line graphs Figure~\ref{fig:combinedacc} reinforce the table's data, showcasing that the Proto-RM model maintains a higher accuracy across epochs compared to the Baseline-RM for the $5\%$, $10\%$, and $20\%$ of the Summarize dataset. Proto-RM not only starts at a higher accuracy but also demonstrates less variability and ends with a higher accuracy, indicating a more robust model.

\begin{table}[!h]
\centering
\small
\begin{tabular}{@{}lccc@{}}
\toprule
& \multicolumn{3}{c}{\textbf{Toxicity}} \\
\cmidrule(l){2-4}
\textbf{RM} & 5\% & 10\% & 15\% \\
\midrule
\textbf{Baseline} & 0.574 & 0.610 & 0.633 \\
\textbf{Proto-RM} & 0.588 & 0.625 & 0.654 \\
\bottomrule
\end{tabular}
\caption{Toxicity levels for the Baseline and Proto-RM.}
\label{tab:toxicity-analysis}
\vspace{-4mm}
\end{table}

\subsection{RLHF Performance}
To ensure the consistency and integrity of the evaluation, we employ GPT-4~\cite{OpenAI_ChatGPT} to assess all outputs from GPT-J (6B)~\cite{gpt-j} across four dimensions.
This methodology aligns with various studies highlighting the capabilities of LLMs to produce high-quality text evaluation that align with or surpass human judgments~\cite{gilardi2023chatgpt,alizadeh2023open,bai2022constitutional}. 
Our scoring criteria encompass factual accuracy, text relevance, information completeness, and clarity of expression, are uniformly applied. 
Each dimension receives a score up to 10, with increments of 0.5. The overall score is derived from the average of four metrics. 

\par \textbf{Accuracy} (Acc): Assesses whether the content of the answer or summary accurately reflects the information and intention of the original prompt.
\par \textbf{Relevance} (Rel): Checks whether the answer or summary is closely related to the original prompt.
\par \textbf{Completeness} (Comp): Evaluates whether the provided information is comprehensive, covering all key points and details in the prompt.
\par \textbf{Expression} (Expr): Considers whether the language expression of the answer or summary is clear and understandable.
\par The results in Figure~\ref{fig:baselineVSours} indicates that the LLM fine-tuned with \model~
outperforms the Baseline across all four aspects, showing an increase from $0.4/10$ to $0.54/10$ in overall score, which is significantly higher than the Baseline. Moreover, it demonstrates a clear advantage in both Accuracy and Expression, with the highest scores reaching $0.76/10$ and $0.82/10$ respectively. Table~\ref{tab:example} demonstrates the differences in the output text quality of GPT-J with no fine-tuning, fine-tuned using Baseline-RM, and fine-tuned using Proto-RM. The discrepancies highlighted also validate the efficacy of our improved reward model.

\begin{table*}
\centering
\small 
\setlength{\tabcolsep}{5pt} 
\begin{tabular}{m{0.25\linewidth} m{0.35\linewidth} m{0.35\linewidth}}
\toprule
\hline
\multicolumn{3}{p{\dimexpr\linewidth-2\tabcolsep\relax}}{\textbf{Prompt:} 
SUBREDDIT: r/relationships
TITLE: My girlfriend (20f) of two years cheated on me (20m) by kissing two guys at a Halloween party.
\par POST: Lately her and I have been having a few problems, and these \colorbox{mybgcolor4}{problems have been brought up before} a few times...
I feel terrible about it, but this time I was really trying to change for her.
For Halloween she went to visit her step brother at a college and I got drunk with my friends and watched movies. Last night (11/1) we got in a huge fight about me not changing and how our relationship won't work out and basically \colorbox{mybgcolor}{broke up} over the phone. So in an effort to try and fix it I drove to her house. She told me how at the parties she went to that \colorbox{mybgcolor3}{two guys kissed} her... Should I even \colorbox{mybgcolor2}{try to fix} it or would I be better off cutting all ties...


TL;DR:
} \\
\hline
\textbf{GPT-J without Fine-tuning} & \textbf{GPT-J with Baseline Reward Model} & \textbf{GPT-J with Protonet-Reward Model} \\
\hline
Girlfriend cheated on me and now we may be fixing things up, but we're not too sure if I should. I love her and we both want to \colorbox{mybgcolor2}{fix things}. Should we? & girlfriend of 2 years \colorbox{mybgcolor3}{kissed two guys} at Halloween party and I don't know what to do. But I want to try and \colorbox{mybgcolor2}{fix it}. & Girlfriend and I \colorbox{mybgcolor}{broke up} over me \colorbox{mybgcolor4}{not changing}, decided to fix that, find out she \colorbox{mybgcolor3}{kissed} someone else at a Halloween party. Should we \colorbox{mybgcolor2}{work it out} or not? \\
\hline
\hline
\end{tabular}
\vspace{-2mm}
\caption{Comparative responses from GPT-J models to a given prompt. The answer from the GPT-J with the Proto-RM is more coherent and relevant. GPT-J with Proto-RM catches all four necessary points in the text, and indicates a higher level of understanding and alignment with human preferences than the other models' responses.
}
\label{tab:example}
\vspace{-4mm}
\end{table*}

\subsection{Ablation Study}
\begin{figure}[h]
  \centering
  \includegraphics[width=0.45\textwidth]{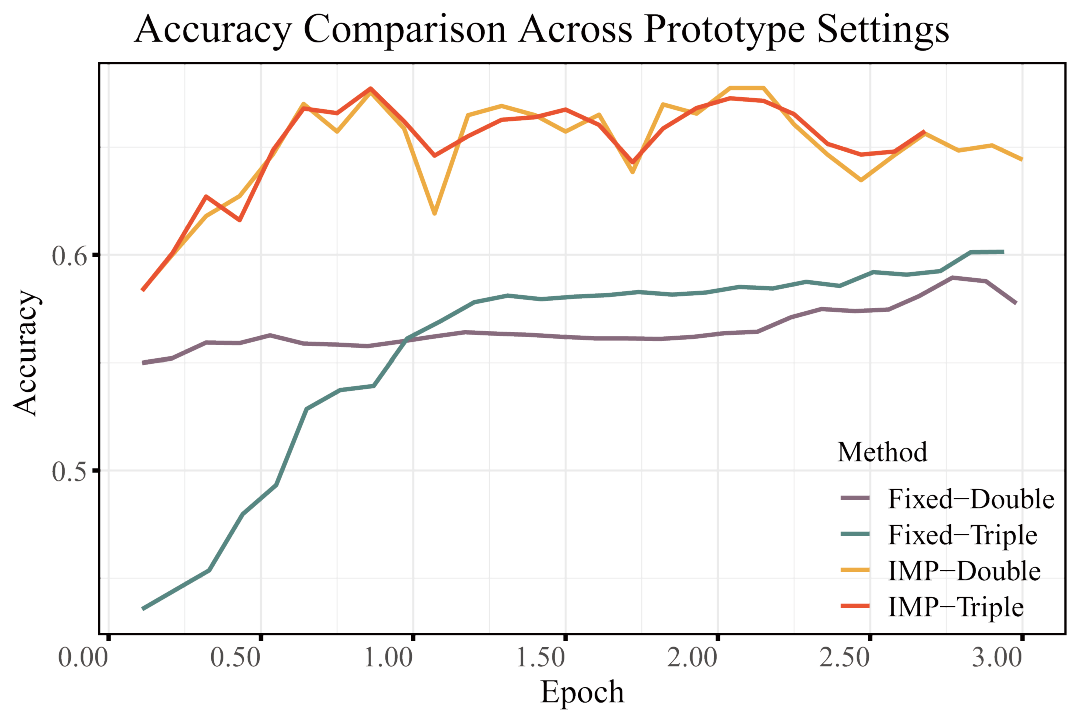}
  \caption{Impact of IMP. Models incorporating IMP demonstrate higher accuracy and more stable accuracy during training.
  }
  \label{fig:protonum}
  \vspace{-4mm}
\end{figure}

\paragraph{Study of IMP.}~\cite{allen2019infinite} We explore and compare the effects of using different numbers of prototypes with various methods for setting the prototype quantities. Specifically, we examine the outcomes of setting the prototype numbers to twice and thrice the default amount and the outcomes of gradually increasing the number of prototypes from the default to double and triple using the IMP method. Figure~\ref{fig:protonum} illustrates that adopting the IMP method for prototype numbers yields better results in both accuracy and stability compared to fixed prototype numbers. The lines representing IMP methods (both IMP-Double and IMP-Triple) show higher accuracy over the epochs. Additionally, the IMP lines demonstrate a smoother progression with less fluctuation, suggesting greater stability in model performance across epochs.

\begin{figure}[h]
  \centering
  \includegraphics[width=0.45\textwidth]{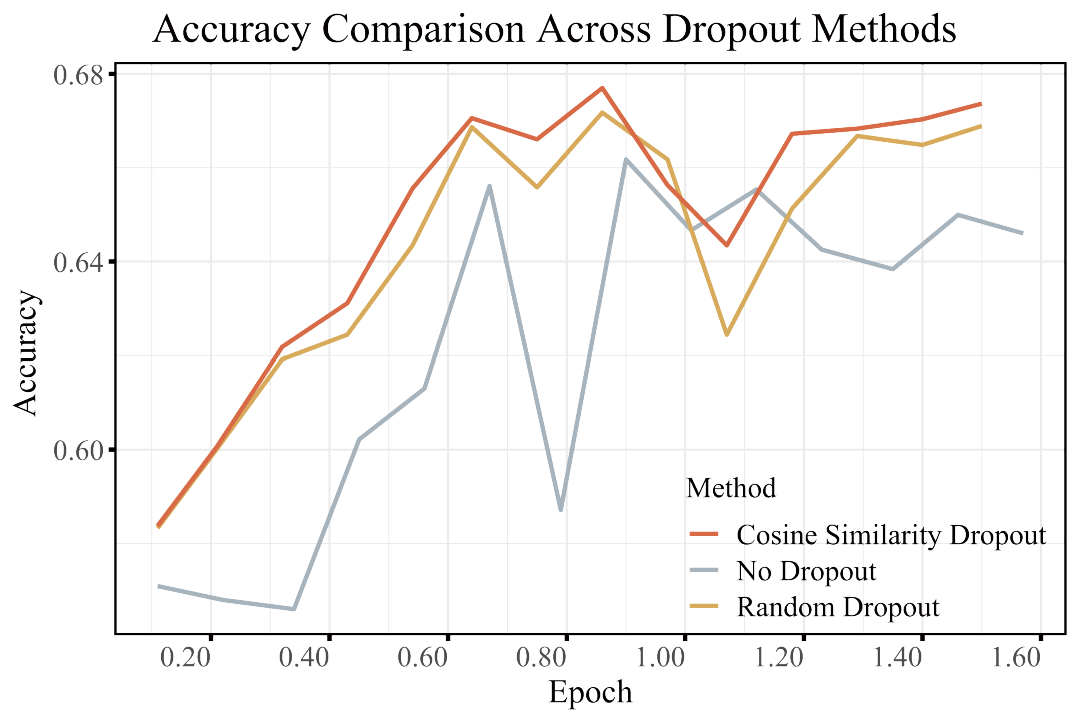}
  \caption{Impacts of Dropout. Models incorporating dropout exhibit higher accuracy, with Cosine Similarity Dropout performing slightly better than Random Dropout.
  }
  \label{fig:dropout}
  \vspace{-4mm}
\end{figure}

\paragraph{Effects of Dropout.} As shown in Figure~\ref{fig:dropout}, we find that employing a Dropout method, which proportionally drops out a part of the prototypes during the sample embedding updates, yields better results. Specifically, as the line chart illustrates, adopting a Dropout method significantly outperforms the approach of not using Dropout in terms of accuracy. Among the Dropout approaches, the method utilizing Cosine Similarity Dropout achieves higher accuracy compared to Random Dropout and exhibits greater stability. This underscores the effectiveness of using Cosine Similarity Dropout. 

\paragraph{Toxicity.} We conduct experiments on different proportions of a toxicity dataset from~\cite{bai2022training} over a single epoch. As shown in Table~\ref{tab:toxicity-analysis}, our Proto-RM method outperforms the Baseline at random data proportions of $5\%$, $10\%$, and $20\%$.

\section{Conclusion}
\vspace{-1mm}

Our research demonstrates the efficacy of prototypical networks in refining RLHF, especially in scenarios with limited human feedback. The enhanced reward model shows a marked improvement in aligning LLM outputs with human preferences, as evidenced by our experimental results. future works can explore the application of our method to more diverse and extensive datasets to further validate its effectiveness and adaptability. 

\clearpage

\paragraph{Limitations}

There are a few limitations to the current framework. First, the Proto-RM performs best for open-domain pairwise human preference tasks that focus on alignment with human judgments. 
It remains unknown if \model yields substantial improvements in scenarios involving non=pairwise tasks or datasets that require less complex decision-making processes. 

Second, we focus our study on the English-language human preference data. This source mainly represents the English-language NLP study. Our study does not include all the necessary research from other nature languages.

We examine how often NLP researchers cite older work by analyzing factors such as the mean age of citations. Our findings indicate a link between these factors and the frequency of citations. However, these links do not prove causation. More studies are needed to understand the reasons behind the citations of older papers.

\paragraph{Ethics Statement}

The Proto-RM framework can enhance the effectiveness and data efficiency of the RLHF process for LLM. However, as the framework uses pre-trained GPT-J as support and GPT-4 as an evaluator, it may inherit the ethical concerns associated with GPT-J and GPT-4, such as responding to harmful queries or exhibiting biased behaviors.

\paragraph{Acknowledgements}
This work was supported by Public Computing Cloud, Renmin University of China, Beijing Outstanding Young Scientist Program (BJJWZYJH012019100020098), Major Innovation \& Planning Interdisciplinary Platform for the ``Double-First Class'' Initiative, Renmin University of China, and fund for building world-class universities (disciplines) of Renmin University of China. This work was partially done at Beijing Key Laboratory of Big Data Management and Analysis Methods, Gaoling School of Artificial Intelligence, Renmin University of China, Beijing 100872, China, Public Policy and Decision-making Research Lab of RUC, Engineering Research Center of Next-Generation Intelligent Search and Recommendation, Ministry of Education, MOE.

\clearpage
\bibliography{custom}
\bibliographystyle{acl_natbib}

\appendix


\newpage

\appendix
\section{Supplementary Statistics and Plots}

\subsection{Data Efficiency}

In this section, we present a comparative analysis of our Proto-RM method against the Baseline results obtained from the standard RLHF on the full-size dataset. Notably, \model demonstrates equivalent effectiveness using only $20\%$ of the data that the Baseline requires $100\%$ to achieve. This significant reduction in data usage without compromising performance underlines the data efficiency of our method.

Here is a detailed table of the performance across various datasets:
\begin{table}[h]
\centering
\begin{tabular}{@{}lcc@{}}
\toprule
Datasets & Baseline 100\% & Proto-RM 20\% \\ \midrule
Webgpt & 60.17\% & 60.56\% \\
Pairwise & 99.89\% & 99.84\% \\
Summarize & 68.88\% & 68.72\% \\ \bottomrule
\end{tabular}
\caption{Performance comparison of Baseline RLHF and Proto-RM method.}
\label{tab:my-table}
\end{table}

For \textit{Webgpt} dataset the Proto-RM method slightly improves the performance, with a performance increase from $60.17\%$ to $60.56\%$. This improvement, while modest, indicates that our approach can enhance model effectiveness even with reduced data input.

For \textit{Pairwise} dataset, the Baseline achieves a near-perfect score of $99.89\%$. The Proto-RM's performance closely follows at $99.84\%$, which is remarkable given that it only uses a fraction of the data.

For \textit{Summarize} dataset, the performance of our Proto-RM method is comparably high at $68.72\%$, closely trailing the Baseline's $68.88\%$. Although there is a minor decrease, it falls within the margin of error and showcases that our approach can maintain high levels of effectiveness in summarizing tasks.

In all instances, the Proto-RM method validates our claim of data efficiency. Our method achieves competitive or even better results than the Baseline with significantly less data. This result shows that Proto-RM is suitable for improving data efficiency when the data cost is high or the data is limited.

\subsection{Human Evaluations}

To avoid any form of tuning specifically tailored to GPT-4's evaluation patterns, we present results in Table~\ref{tab:human-base} and Table~\ref{tab:human-proto} evaluated by humans, which may slightly vary from GPT-4's, but also validate the effectiveness of our method in aligning with human preferences.

\subsection{Time Complexity}

Here we analyze the time cost of Proto-RM. For all samples within each minibatch, the prototype updates introduce a time complexity of $O(n)$, where $n$ is the number of samples. For each sample, Proto-RM needs to identify the closest $k$ prototypes out of all $m$ prototypes to compute the output of the prototype network. Although the IMP method gradually increases the number of prototypes, we have set a cap on the maximum number of prototypes that can be added. This cap is defined as $\alpha$ times the initial number of prototypes, where $\alpha$ is a fixed constant. Here, computing the output of the prototype network for each sample incurs a time complexity of $O(km)$. Thus, the introduction of prototype networks results in a time complexity of $O(nkm)$.

This analysis indicates that including prototypical networks does not significantly elevate the overall complexity. Additionally, we provide an analysis of the time taken for training and testing on a dataset of 8,552 samples to support our assertion that Proto-RM does not considerably increase computational costs in Table~\ref{tab:time-analysis}. We believe that the added complexity is manageable and justified by the performance improvements, especially considering the high data costs in such scenarios.

\begin{table}
\centering
\begin{tabular}{@{}lc@{}}
\toprule
Method & Time \\ \midrule
Baseline & 1:32:11 \\
Proto-RM & 1:58:08 \\ \bottomrule
\end{tabular}
\caption{Training and testing time for the Baseline and Proto-RM methods.}
\label{tab:time-analysis}
\end{table}

\begin{table*}
\centering
\begin{tabular}{@{}lcccccc@{}}
\toprule
Metrics & Human 1 & Human 2 & Human 3 & Human 4 & Human 5 & GPT-4 \\ \midrule
Accuracy & 6.61 & 6.56 & 6.52 & 6.82 & 7.14 & 6.73 \\
Relevance & 7.17 & 7.33 & 6.98 & 7.15 & 6.90 & 7.3 \\
Completeness & 6.41 & 7.10 & 6.79 & 6.42 & 7.28 & 6.50 \\
Expression & 7.33 & 7.33 & 6.99 & 7.45 & 7.40 & 7.24 \\
Overall & 6.88 & 7.08 & 6.82 & 6.96 & 7.18 & 6.94 \\ \bottomrule
\end{tabular}
\caption{Human Annotations on Summarize from Feedback for Baseline Model.}
\label{tab:human-base}
\end{table*}

\begin{table*}
\centering
\begin{tabular}{@{}lcccccc@{}}
\toprule
Metrics & Human 1 & Human 2 & Human 3 & Human 4 & Human 5 & GPT-4 \\ \midrule
Accuracy & 7.02 & 7.28 & 6.80 & 7.21 & 7.34 & 7.07 \\
Relevance & 7.11 & 7.35 & 7.35 & 7.41 & 7.02 & 7.40 \\
Completeness & 6.62 & 7.24 & 7.08 & 6.71 & 7.71 & 6.93 \\
Expression & 7.33 & 7.35 & 7.49 & 7.47 & 7.69 & 7.57 \\
Overall & 7.02 & 7.31 & 7.18 & 7.20 & 7.44 & 7.24 \\ \bottomrule
\end{tabular}
\caption{Human Annotations on Summarize from Feedback for Proto-RM model.}
\label{tab:human-proto}
\end{table*}

\end{document}